\documentclass[]{spie}  

\usepackage{amsmath,amsfonts,amssymb}
\usepackage{graphicx}
\usepackage{multirow} 
\usepackage[colorlinks=true, allcolors=blue]{hyperref}

\title{Two-stage iterative Procrustes match algorithm and its application for VQ-based speaker verification}

\author[a]{Richeng Tan}
\author[b]{Jing Li}
\affil[a]{Baidu Inc, tanricheng@baidu.com, Beijing, China}
\affil[b]{GAC Automotive Engineering Institute,lijing2@gaei.cn,Guangzhou, China}

\authorinfo{Further author information: (Send correspondence to Richeng Tan)\\Richeng Tan: E-mail: tanricheng@baidu.com, Telephone: +86 15116240052\\  Jing Li: E-mail: lijing2@gaei.cn }

\begin{document} 
\maketitle

\begin{abstract}
In the past decades, Vector Quantization (VQ) model has been very popular across different pattern recognition areas, especially for feature-based tasks. However, the classification or regression performance of VQ-based systems always confronts the feature mismatch problem, which will heavily affect the performance of them. In this paper, we propose a two-stage iterative Procrustes algorithm (TIPM) to address the feature mismatch problem for VQ-based applications. At the first stage, the algorithm will remove mismatched feature vector pairs for a pair of input feature sets. Then, the second stage will collect those correct matched feature pairs that were discarded during the first stage. To evaluate the effectiveness of the proposed TIPM algorithm, speaker verification is used as the case study in this paper. The experiments were conducted on the TIMIT database and the results show that TIPM can improve VQ-based speaker verification performance clean condition and all noisy conditions. 
\end{abstract}

\keywords{iterative Procrustes matching, frame selection, speaker verification, Vector Quantization}

\section{INTRODUCTION}
\label{sec:intro}  

Vector Quantization (VQ) is a classical quantization technique that allows the modeling of probability density functions by the distribution of some prototype vectors. VQ-based systems first cluster the original data, and then use the center value of each cluster to represent them. In the past decades, VQ has been very popular in the field of speaker recognition due to its fast processing speed and good recognition performance. Although Gaussian Mixture Model \cite{reynolds1995robust} and i-vector \cite{dehak2011front} have shown better performance for speaker-related task in recent years, VQ-based systems are more suitable for the application that only have a small amount of training data in comparison to them. Matsui et al.\cite{matsui1994comparison}  compared VQ based system with the Hidden Markov Models (HMM) based speaker identification system and the result shows that VQ based system is more robust than the HMM based system when there is only a small amount of training data. 

Unfortunately, in real-world application, the background noise, channel mismatch and many other factors could cause the feature mismatching problem and reduce the performance of VQ-based model, which would heavily degrade the performance of the speaker verification performance. To solve this problem, speech enhancement \cite{wong2001text,zhao2014robust,sagayama1997jacobian} and model adaptation approaches \cite{xue2016speaker,xue2014direct} have been widely used. Since the same noise may pose different distortion on different frames, by adopting several noisy constraints, Song et al.\cite{song2018noise} designed a simple framework that can select noise invariant frames from original audio signals. The experiment results show that this framework can enhance the speaker verification performance for different speaker models under different noisy conditions. 
As for feature mismatching problem, Random Sample Consensus (RANSAC) \cite{fischler1987random} has been proved to be a very efficient method for eliminate mismatched feature points and has been applied to many image-based tasks, for example, image retrieval \cite{song2015precise}, object recognition \cite{li2018multi} etc. However, RANSAC utilizes the geometric information of matched image feature points which audio feature does not have. Meanwhile, Procruste-based method can also eliminate the mismatched feature points in image matching and the geometric information is not necessary for this method. Therefore, it is suitable for audio-based feature matching tasks. This paper proposes a two-stage iterative Procrustes match (TIPM) approach aiming to remove the mismatched feature vectors for VQ-based speaker verification. At the first stage, TIPM will remove those mismatched feature vectors pairs and recycle incorrectly removed feature vectors pairs at the second stage.

To evaluate the performance of the proposed method, we conducted two experiments on TIMIT \cite{zue1990speech} database by comparing the VQ-based speaker verification performance with and without using TIPM (illustrated in Figure 1). 

   \begin{figure} [ht]
   \begin{center}
   \begin{tabular}{c} 
   \includegraphics[height=5.3cm]{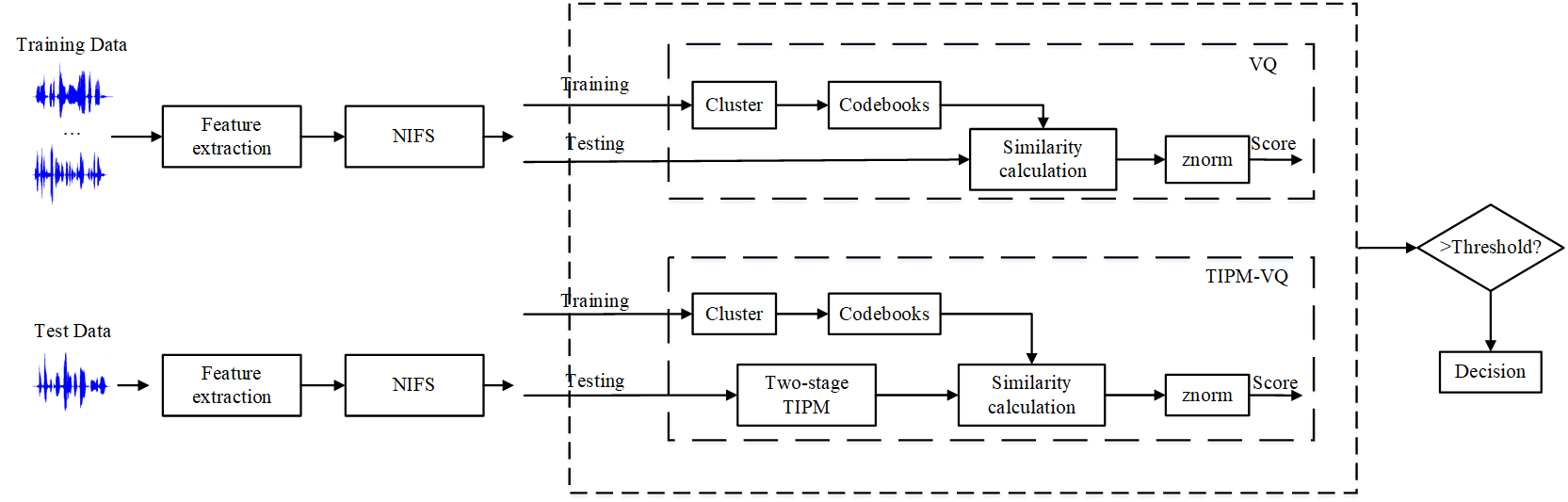}
	\end{tabular}
	\end{center}
   \caption[example] 
   { \label{fig:1} System flow chart}
   \end{figure} 
The rest of this paper is structured as follows. Our detailed TIPM method is explained in Section 2. Experimental results are discussed in Section 4. The last section is devoted to the conclusion.

\section{Methodology}
\label{section2}
For VQ-based speaker verification, the vector set $\overline{\overline{F}}$  obtained from a speaker can be grouped into several subsets, while each of them contains several feature vectors that are similar and may represent the same character of a speaker. Meanwhile, the matching speed would be heavily affected if there are a large number of feature vectors that need to be processed. Motivated by this, K-means is introduced to cluster feature vectors of each speaker. As a result, the final codebooks of all speakers can be generated, which are defined as 
	   $$C_n=\{C^n_1,C^n_2,L,C^n_{Q^n}\}\eqno(1)$$	
where $N$ is the number of codebooks and $C^n_q,q=1,2,L,N$;$Q^n$is the number of clusters of codebook n；$C^n_q,q=1,2,L,Q^n$   are the vectors of each cluster center of the codebook n.
The matching score of the test trial determines whether an utterance would be accepted in a speaker verification system. Conventionally, the score is computed by the Euclidean metric or log likelihood rate. Although the NIFS scheme attempt to select the noise robustness frames, additive noises may still impose relative distortions on the selected frames. This would lead many incorrect matching. The Procrustes-based feature match used in \cite{sheng2010two} is one of the typical approaches to solve this problem. It takes the geometric distribution of feature vectors into consideration, which is expected to help differentiate speakers. In this section, a two-stage iterative Procrustes matching algorithm (TIPM) is proposed, which aims to minimize the number of mismatched feature vector pairs.
\par Assume that $C_a=\{C^a_1,C^a_2,L,C^a_{Q^n}\}$  is the codebook of a speaker A and $\overline{{F_{test}}}=\{\overline{f^{test}_1},\overline{f^{test}_2},L,\overline{f^{test}_3}\}$  is the NIFV of a test utterance, which selected by the first-stage NIFS, then a set of initial matching feature vector pairs could be obtained according to
	    	$$d(C^a_q,\overline{f^{test}_p})\leq\epsilon_t\eqno(2)$$
\par where $\overline{f^{test}_p}$ is the nearest neighborhood feature vector of $C^a_q$; is a threshold of Euclidean distance and $q=1,2,L,Q^a,p=1,2,L,P$. Assume that the matrix $X_l$ is established by the matched feature vectors from the codebook $C_a$, while their corresponding feature vectors from the test utterance $\overline{F_{test}}$  established the matrix $X_q$, based on the equation (14) and (15) the orthogonal matrix  could be constructed.	
	   	$$\Gamma=argmax\|X_j{\Omega}X_q\|\eqno(3)$$
	 	  $${\Omega}^T{\Omega}=I\eqno(4)$$
\par where $X_l=\{x_{l_s},s=1,2,...,S\}$ and $X_q=\{x_{q_s},s=1,2,L,S\}$  ; S is the number of original matched pairs. The F-norm   is minimized by the nearest orthogonal matrix, and
	  	  $${\Gamma}={\Psi}{\Upsilon}^*\eqno(5)$$
	 	  $${X^T_l}{X_q}={\Psi}{\sum}{\Upsilon}\eqno(6)$$
\par where ${\Psi}{\sum}{\Upsilon}$  is the singular value decomposition (SVD) of ${X^T_l}{X_q}$  . We define $e(X_l,X_q)$  as the similarity measurement between  $X_l$  and $X_q$  , which is the least square error of  $X_l$  and $X_q$    in Procrustes match and the best case is $e(X_l,X_q)=0$  . Suppose that there does exist some mismatched vector pairs, then the value of $e(X_l,X_q)$   would above zero. If a pair of matched vectors $X^i_l$  in $X_l$  and $X^i_q$ in $X_q$ are discarded, the process is called  $X_{l{\leftarrow}q}$  and $X_{q{\leftarrow}i}$  . The vector pair $X^i_l$   and $X^i_q$   would be discarded from (7)  and (8) , if
$$e(X_{l{\rightarrow}i},X_{q{\rightarrow}i})/e(X_l,X_q)<\delta\eqno(7)$$
$$e(X_{l{\rightarrow}i},X_{q{\rightarrow}i})=min\{e(X_{l{\rightarrow}l_s},X_{q{\rightarrow}}q_s)\}\eqno(8)$$
where  $\delta$  is a iteration threshold and $s=1,2,...,S$  . As a result, a new pair of matrixes $X_{l{\rightarrow}i}$   and $X_{q{\rightarrow}i}$   could be obtained. Consequently, a new least square error  $e(X_{l{\rightarrow}i},X_{q{\rightarrow}i})$  is generated. Afterwards, by repeating the procedure mentioned above, another pair of feature vectors  $X^j_{l{\rightarrow}i}$  and $X^j_{l{\rightarrow}i}$   may be removed from  $X_{l{\rightarrow}i}$  and  $X_{l{\rightarrow}i}$ . This leave-one-out procedure can proceed iteratively until equation (7) and (8) are no longer satisfied. Finally, n pairs of mismatched vectors are discarded from  $X^i_l$  and $X^i_q$  , which can be denoted as
	  	 $$dis(p)=\{p_i,i=1,2,L,n\},n<S$$
Then, a pair of new matrix  $Y_l$  and  $Y_q$  are generated.
However, some vector pairs may be removed incorrectly at the first stage. To recycle them, the second-stage of the Procrustes match is introduced. Suppose that the least square error of $Y_l$   and  $Y_q$  is $e(Y_l,Y_q)$  . We define that if a pair of points  $Y^i_l$  and $Y^i_q$  is added from $dis(p)$  to $Y_l$   and $Y_q$  , a new pair of matrices  $Y{l{\leftarrow}t}$ and $Y{q{\leftarrow}t}$ would be generated. Then, by adding each pair of feature vectors in the  $dis(p)$  to $Y_l$   and $Y_q$  respectively, a set of $e(X_{l{\rightarrow}i},X_{q{\rightarrow}i}),i=1,2,L,n$  could be obtained. A pair of vectors  $Y^i_l$  and $Y^i_q$  would be recycled from $dis(p)$  and added to $Y_l$   and  $Y_q$  if
$$e(Y_{l{\rightarrow}i},Y_{q{\rightarrow}i})/e(Y_l,Y_q)<\eta\eqno(9)$$
$$e(X_{l{\rightarrow}i},X_{q{\rightarrow}i})=min\{e(Y_{l{\rightarrow}l_s},Y_{q{\rightarrow}}q_s)\}\eqno(10)$$
where  $\eta$  is a iteration threshold and $i=1,2,L,n$ . Afterwards, by repeating the procedure mentioned above, a new pair of feature vectors  $Y^j_{l{\leftarrow}i}$  and $Y^j_{q{\leftarrow}i}$    may be recycled from $dis(p)$  and added to $X_{l{\Leftarrow}i}$   and  $X_{q{\Leftarrow}i}$ . This add-one-in procedure can proceed iteratively until equation (9) and (10) are no longer satisfied. 
Consequently, the final matched matrices  $Z_l$  and $Z_q$  are obtained, which contain the final matched vector pairs. (Figure 2. a)  illustrates the relationship among different kinds of matched pairs. 

\begin{figure*}[!htbp]
   \centering
   \includegraphics[width=16cm,height=8cm]{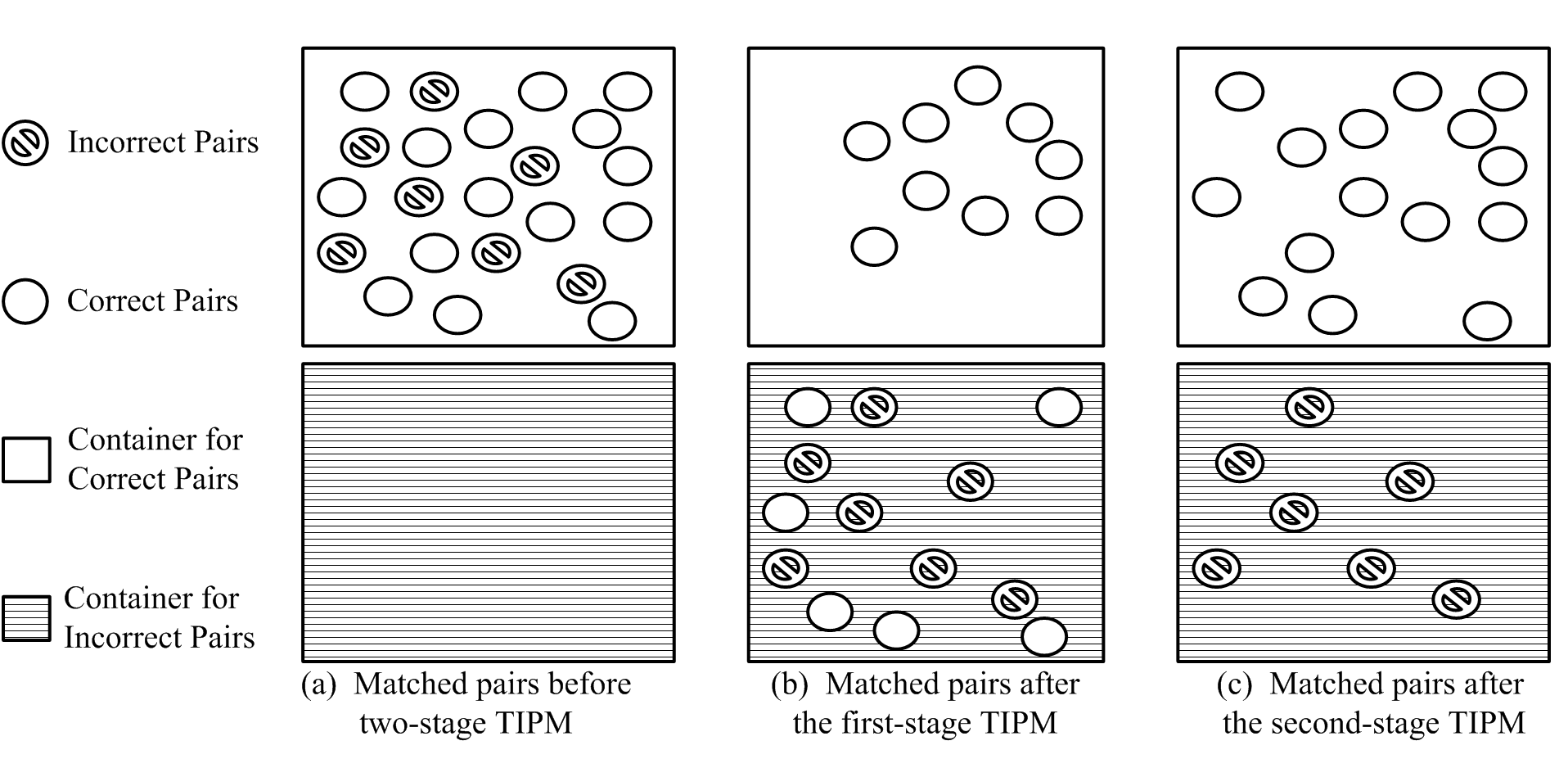}
    \caption{The principle of the TIPM}
    \label{fig.1}
\end{figure*}

It demonstrates that initial matched pairs are made up of mismatched pairs and the final matched pairs. (Figure 2) shows the change of matched vector pairs during TIPM.

Since the number of frames selected from each test utterance may differ, a relative metric is adopted for VQ-based systems (VQ baseline, NIFS-VQ, NIFS-TIPM-VQ) to measure the similarity between a codebook and a test utterance, given by
$$L=\frac{N(M(\overline{F_{test}},C_a))}{N(\overline{F_{test}})}\eqno(11)$$
where $N(M(\overline{F_{test}},C_a))$   is the number of the final matched vector pairs between the test utterance and the codebook, and  $N(\overline{F_{test}})$  is the number of selected frames in $\overline{F_{test}}$  . According to the equation(11), the larger number of matched vectors the test utterance has with a codebook, the higher similarity ratio between them. In order to minimize the inconsistency of feature vectors from the same speaker and maximize the inconsistency of feature vectors from different speakers, the similarity score would be further normalized by zero normalization according to
$$L^{\star}(V)=\frac{L(V)-\mu}{\eta}\eqno(12)$$
where  $L(V)$  is the original score of sample $V$  ; $\mu$  and $\eta$   are the estimated impostor parameters for speaker model; $L^{\star}(V)$  is the score after zero normalization .

\section{EXPERIMENT RESULT}
\label{section3}

	\subsection{Experiment setup}
To evaluate the usefulness of the proposed TIPM, we conducted two experiments on the TIMIT database. The first one is applying TIPM directly to the feature vectors of each voice sample while the second one is to apply TIPM to the feature vectors that are already processed by the NIFS. Besides the clean condition, in order to test the performance of TIPM under noisy conditions, 12 noisy conditions consisting of four different noises with three different SNRs. Please see for details of the method for creating these noisy conditions.  
The measurement utilized here to evaluate the performance of the usefulness of TIPM are the Equal Error Rate (EER) values and DET curves.

	\subsection{Experiment result}
We first apply TIPM to the traditional VQ system. Table compares the speaker verification result of the TIPM-VQ system with it of the traditional VQ system. It is clear that the TIPM has improved the performance in clean environment and 10 noisy environments. After using TIPM, the performance falling down in only two noisy conditions with SNR of 15. Particularly, when applied TIPM under the clean and noisy conditions with high SNR (25), it achieved promising result in terms of relative improvement of EER. 

Figure 3 and Table 1 demonstrate the DET and EER results between NIFS-VQ system and TIPM-NIFS-VQ system. It could be noted from the DET curves that by applying TIPM algorithm to the NIFS-VQ system, the performances under all conditions have been further improved. Meanwhile, When TIPM-NIFS-VQ system operated in all 12 noisy environments, the EER values dropped under all conditions except the volvo noisy environment with SNR of 25dB, where it kept stable at 11.88\%, in comparison with NIFS-VQ system. Meanwhile, NIFS-TIPM-VQ system obtained the best EER result with 7.12\% among all three VQ-based systems in the clean environment.

\begin{figure*}[!htbp]
   \centering
   \includegraphics[width=18cm,height=13cm]{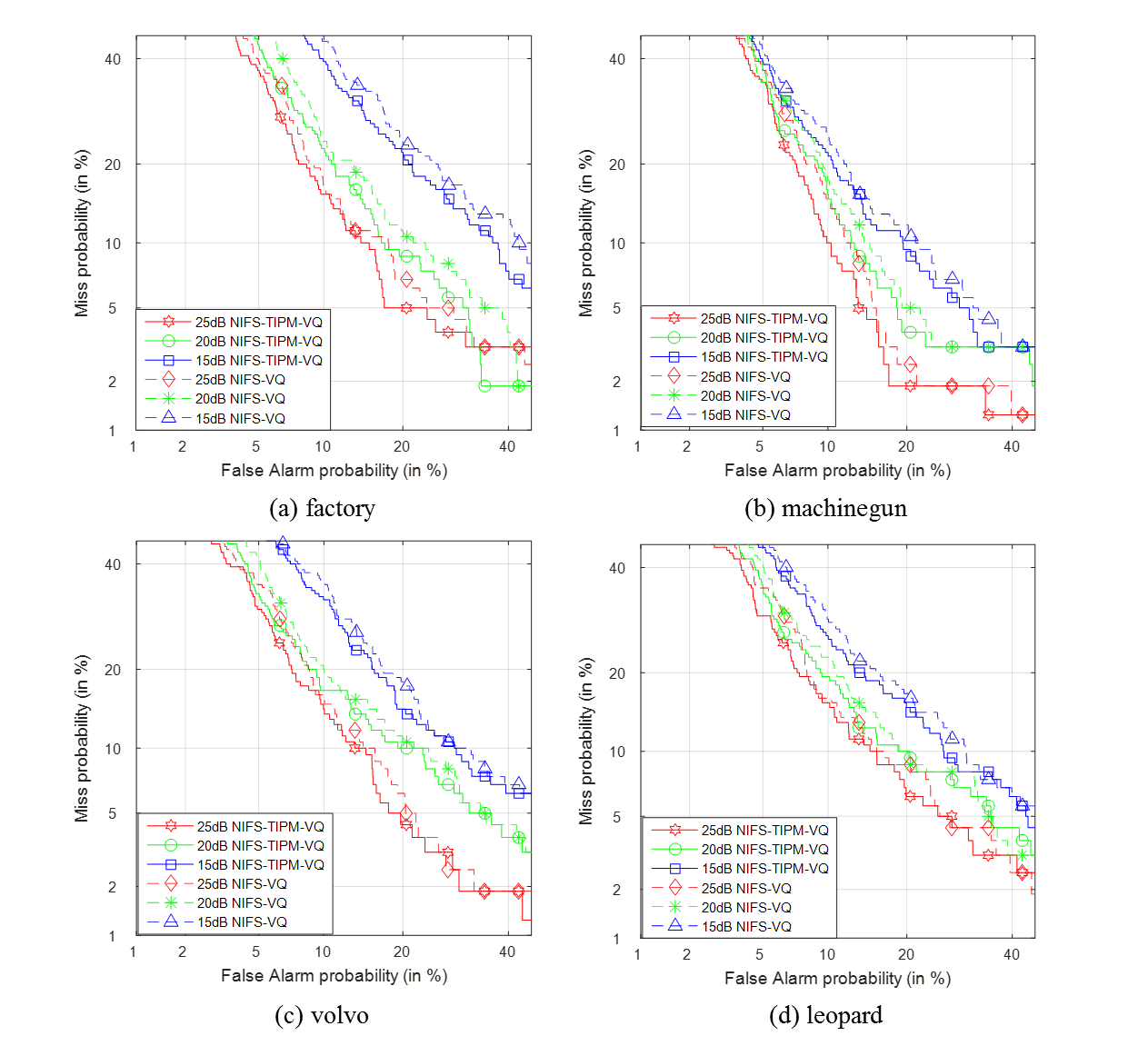}
    \caption{Experiment results}
    \label{fig.3}
\end{figure*}

\begin{table*}[!htbp]
\centering
\caption{THE EER OF VQ AND TIPM-VQ}\label{tab:aStrangeTable}
\begin{tabular}{cccccc}
\hline
& SNR& VQ baseline&TIPM-VQ & Absolute improvement &Relative improvement(\%)\\
\hline
clean& &16.9 &14.6 &2.3&8.6\\
\hline
\multirow{3}*{factory}&25&23.1&20.1&3.0&12.9 \\
&20&26.9&25.3&1.6&5.9 \\
&15&32.8&33.6&-0.8&-2.4 \\
\hline
\multirow{3}*{matchinegun}&25&18.1&17.7&0.4&2.2 \\
&20&20.4&19.1&1.3&6.4 \\
&15&24.5&23.6&0.9&3.7 \\
\hline
\multirow{3}*{volvo}&25&18.2&16.9&1.3&7.1 \\
&20&20.6&20.3&0.3&1.5 \\
&15&26.5&25.2&1.3&4.9 \\
\hline
\multirow{3}*{leopard}&25&23.1&20.4&2.7&11.7 \\
&20&26.9&26.0&0.9&3.3 \\
&15&32.7&33.1&-0.4&-1.2 \\
\hline
\end{tabular}
\end{table*}

\begin{table*}[!htbp]
\centering
\caption{THE EER OF NIFS-VQ AND NIFS-TIPM-VQ}\label{tab:aStrangeTable}
\begin{tabular}{cccccc}
\hline
& SNR& NIFS-VQ &NIFS-TIPM-VQ & Absolute improvement &Relative improvement(\%)\\
\hline
clean& &8.34 &7.12 &1.22&14.63\\
\hline
\multirow{3}*{factory}&25&12.50&12.19&0.31&2.48 \\
&20&15.63&14.38&1.25&8.00 \\
&15&22.47&20.63&1.84&8.19 \\
\hline
\multirow{3}*{matchinegun}&25&11.61&10.00&1.61&13.88 \\
&20&12.91&11.76&1.15&8.91 \\
&15&14.66&13.86&0.80&5.46 \\
\hline
\multirow{3}*{volvo}&25&11.88&11.88&0.00&0.00 \\
&20&15.00&13.75&1.25&8.33 \\
&15&18.75&17.63&1.12&5.97 \\
\hline
\multirow{3}*{leopard}&25&13.13&12.19&0.94&7.16 \\
&20&14.38&13.19&1.19&8.28 \\
&15&18.34&16.88&1.46&7.96 \\
\hline
\end{tabular}
\end{table*}
	\subsection{Result analysis}
The feature matching algorithm (TIPM） proposed in this paper can influence the number of matched feature vector pairs employed from the total available original matched feature vectors. When applied to the traditional VQ system, after the first-stage Procrustes match, the average matched pairs was decreased from 65.2\% to 46.3\% and from 14.7\% to 9.9\% for NIFS-VQ system. This is because the mismatched feature vector pairs were discarded. The figures increased to 51.2\% and 11.5\% for each system respectively after the second-stage TIPM, which revealed that some incorrectly discarded feature vector pairs have been recycled. In other words, the second-stage of the algorithm recycled some matched feature vector pairs that had been incorrectly discarded in the first-stage TIPM. In terms of the speaker verification result, it is clear that the proposed TIPM algorithm is effective when combined with either a regular speaker model (VQ) or combined with another preprocessing method (NIFS) with speaker model.
	
\section{CONCLUSION AND DISCUSSION}
\label{section4}
For feature matching tasks, the incorrect match in the feature matching step always pose an negative impact on the performance. Aiming to solve this problem, this paper proposed a two-stage iterative Procrustes match algorithm that can discard mismatched feature vectors pairs between test data and codebooks in VQ-based systems. In order to evaluate the usefulness of the algorithm, speaker verification is used for case study. Two experiments were conducted by VQ baseline VS TIPM-VQ and NIFS-VQ VS NIFS-TIPM-VQ on a subset of the TIMIT database. Besides the clean condition, 12 different noisy conditions were also introduced, which are made up of four different noises with three different SNRs. The experiment result proved that the EER under almost all conditions can be slightly improved by using the TIPM. In addition, it even can obtain better result in 10 out of 12 environments when cooperated with a previous pre-processing method, which illustrated that it has a potential to work with other pre-processing or post-processing methods.

\bibliography{report} 
\bibliographystyle{spiebib} 

\end{document}